\def\year{2019}\relax
\documentclass[letterpaper]{article} 
\usepackage{aaai19}  
\usepackage{times}  
\usepackage{helvet}  
\usepackage{courier}  
\usepackage{url}  
\usepackage{graphicx}  
\usepackage{algorithm2e}
\usepackage{amsmath}
\usepackage{xspace}
\usepackage{dialogue}
\newcommand{\spire}{{\sc Spire}\xspace}
\newcommand{\citenopar}[1]{\citeauthor{#1} \citeyear{#1}}

\begin{document}
%

\title{Combining Deep Learning and Qualitative Spatial Reasoning to Learn Complex Structures from Sparse Examples with Noise}
\author{Nikhil Krishnaswamy$^{1}$, Scott Friedman$^{2}$ and James Pustejovsky$^{1}$ \\
$^{1}$Brandeis University Dept. of Computer Science, Waltham, MA, USA\\
$^{2}$Smart Information Flow Technologies, Minneapolis, MN, USA\\
{\tt nkrishna@brandeis.edu, friedman@sift.net, jamesp@brandeis.edu}\\
}
\maketitle
\begin{abstract}
Many modern machine learning approaches require vast amounts of training data to learn new concepts; conversely, human learning often requires few examples---sometimes only one---from which the learner can abstract  structural concepts.  We present a novel approach to introducing new spatial structures to an AI agent, combining deep learning over qualitative spatial relations with various heuristic search algorithms.  The agent extracts spatial relations from a sparse set of noisy examples of block-based structures, and trains convolutional and sequential models of those relation sets.  To create novel examples of similar structures, the agent begins placing blocks on a virtual table, uses a CNN to predict the most similar complete example structure after each placement, an LSTM to predict the most likely set of remaining moves needed to complete it, and recommends one using heuristic search.  We verify that the agent learned the concept by observing its virtual block-building activities, wherein it ranks each potential subsequent action toward building its learned concept.  We empirically assess this approach with human participants' ratings of the block structures.  Initial results and qualitative evaluations of structures generated by the trained agent show where it has generalized concepts from the training data, which heuristics perform best within the search space, and how we might improve learning and execution.

\end{abstract}
\section{Introduction}
A distinguishing factor of general human intelligence is the capacity to learn new concepts from abstractions and few examples, either by composing a new concept from primitives, or relating it to an existing concept, its constituent primitives, and the constraints involved \cite{gergely2002developmental}.  Recent research in artificial intelligence has pursued ``one-shot learning," but the prevailing machine learning paradigm is to train a model over a number of samples and allow the AI to infer generalizations and solutions from the model.  This approach is often very successful, but often requires large amounts of data and fails to transfer task knowledge between concepts or domains.

For a computational learner, a complex building action might be composed of moving objects into a specific configuration. The class {\it move} is taken as primitive, being a superclass of {\it translate} and {\it rotate}, the two types of motions that any rigid body can undertake in 3D space.  Thus, {\it move}, {\it translate}, and {\it rotate} can be composed into more complex actions, which define a compostional higher-level concept, such as building a structure, which can then be labeled \cite{langley2006unified,laird2012soar,menager2016episodic}.

However, multiple paths to the desired goal often exist. Structural components may be interchangeable and the order in which relations between component parts can be instantiated is often non-deterministic, especially in the early steps, simply due to there being many ways of solving a given problem, and many ways to generalize from an example.  Computational approaches may handle this class of problem either heuristically (cf. classic A* pathfinding \cite{hart1968formal}) or through reinforcement learning \cite{asada1999cooperative,smart2002effective,williams1992simple} and policy gradients to shrink the search space \cite{gullapalli1990stochastic,peters2008reinforcement}.  

Here, we define a means to use deep learning in a larger learning and inference framework over few samples, in a search space where every combination of configurations may be intractable.  Three-dimensional environments provide a setting to examine these questions in real time, as they can easily supply both information about relations between objects and naturalistic simulated data.  Fine-grained 3D coordinates can be translated into qualitative relations that allow inference over smaller datasets than those required by continuous value sets.  The aforementioned motion primitives can be composed with sets of spatial relations between object pairs, and machine learning can attempt to abstract the set of primitives that hold over most observed examples.

\section{Related Work}
Learning definitions of primitives has long been an area of study in machine learning \cite{quinlan1990learning}. Confronting a new class of problem by drawing on similar examples renders the task decidable, as does the ability to break down a complex task into simpler ones \cite{veeraraghavan2007learning,dubba2015learning,wu2015watch,alayrac2016unsupervised,fernando2017unsupervised}.  Recommendation systems at large propose future choices based on previous ones, which can be considered special cases of ``moves" from one situation to another \cite{smyth2007case}.  Often an example must be adapted to a new situation of identifiable but low similarity, and this knowledge adaptation has also made use of recent advances in machine learning \cite{craw2006learning}.

There has also been much work in extracting primitives and spatial relations from language (e.g., \citenopar{kordjamshidi2011relational}) and images (e.g., \citenopar{muggleton2017meta}; \citenopar{binong2018extracting}; \citenopar{liang2018visual}) and in performing inference from this extracted information (e.g., \citenopar{barbu2012simultaneous}; \citenopar{das2017embodied}).  An agent must be able to unify various ways of describing or depicting things, and then assign a label to those novel definitions \cite{hermann2017grounded,narayan2017towards,al2017natural}.  Given a label and known examples, an open question in cognitive systems research concerns the ability of an agent to generate through analogy novel examples of the objects or actions falling under said label, within variation allowed by an open world \cite{r3_acs_2017,alomari2017learning}.


\section{Data Gathering}
To examine this, we used data from a study where users collaborated with a virtual avatar to build 3-step, 6-block staircases out of uniquely-colored virtual blocks on a table in a virtual  world \cite{krishnaswamy2018LREC}.\footnote{https://github.com/VoxML/public-data/tree/master/AvatarInteraction/PilotStudy-2018Feb}

Users interacted with an avatar implemented in the VoxSim experimental environment \cite{krishnaswamy2016multimodal,krishnaswamy2016voxsim} and VoxML platform \cite{pustejovsky2016LREC}.  Using natural language and gestures recognized from depth data by deep convolutional neural networks (DCNNs) in real time, users directed the avatar until they reached a structure that satisfied their definition of a 3-step staircase.  Placing even a simple blocks world scenario in a 3D environment opens the search space to the all the complex variation in the environment, meaning that an enormous number of sets of relative offsets and rotations may define structures of the same label.  Hence building fairly simple structures can have exponential search space.

\begin{figure}
\centering
\includegraphics[height=.6in]{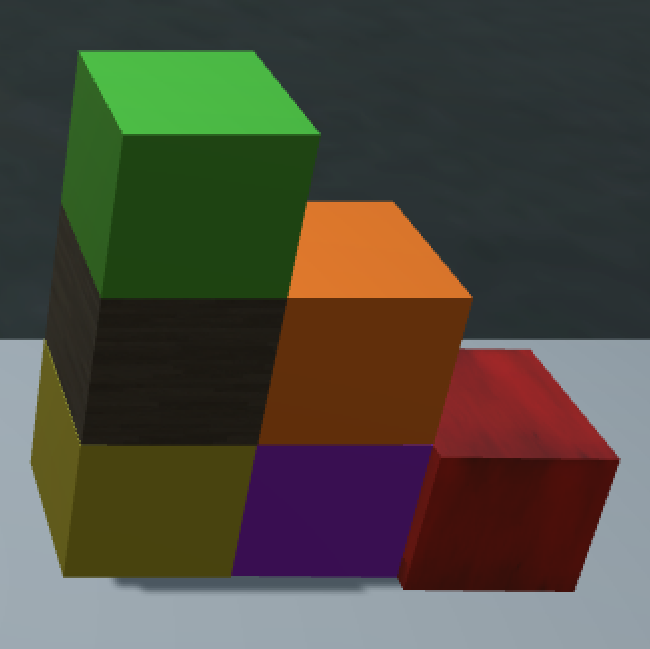}
\includegraphics[height=.6in]{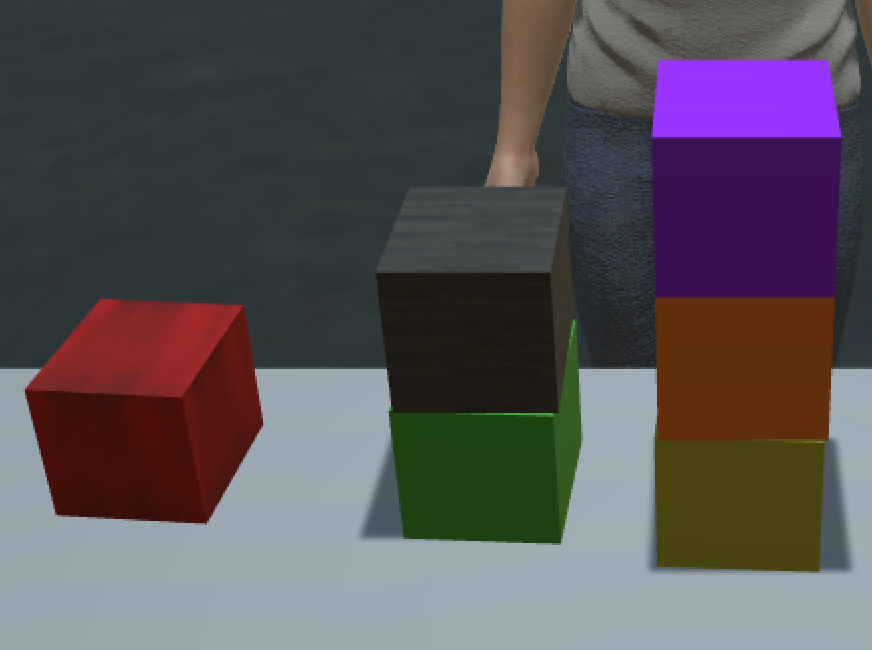}
\includegraphics[height=.6in]{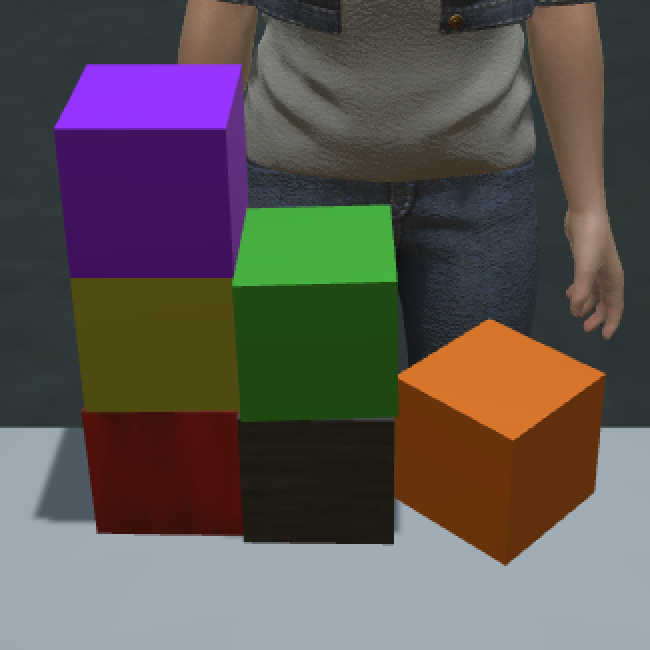}
\caption{\label{fig:examples}Example user-constructed staircases}
\end{figure}

The data consists of log files recording each session of user-avatar interaction which, when rerun through the VoxSim system, recreates the structure built by that user in that session.  We recreated 17 individual staircases which range from highly canonical to highly variant in terms of spaces between blocks or rotated blocks (Fig.~\ref{fig:examples}).  In all structures, the particular placement of individual blocks varied, as a block of any color could be placed anywhere in the staircase.  We then extracted the qualitative relations that existed between all blocks on the table as defined in a subset of the Region Connection Calculus (RCC8) \cite{randell1992} and Ternary Point Configuration Calculus (TPCC) \cite{moratz2002qualitative} included in QSRLib \cite{gatsoulis2016qsrlib}.  Where QSRLib calculi only cover 2D relations, VoxSim uses extensions such as RCC-3D \cite{albath2010rcc} or computes axial overlap with the Separating Hyperplane Theorem \cite{schneider2014convex}.  The set of relations defining each structure is stored as an entry in a database (example shown in Table~\ref{fig:examples}).  Over the dataset, structures were defined by approximately 20 block-block relations.

The relation vocabulary extracted from the examples was {\tt left}, {\tt right}, {\tt touching}, {\tt under}, and {\tt support}.\footnote{These are convenience labels for relations or combinations of relations denoted otherwise in QSRLib.  For example {\it touching} is equivalent to {\it externally connected} ($EC$) in RCC or RCC-3D.}  These combine under certain conditions, such as {\tt left,touching} or {\tt under,touching,support}.  {\tt support} encodes only direct support, such that if block $x$ supports block $y$ and block $y$ supports block $z$, block $x$ would not {\tt support} block $z$ ($x$ would be {\tt under} $z$, but not {\tt touching}---thus {\tt under,touching,support} is considered the inverse of {\it on}).  {\tt left} and {\tt right} are paired such that $left(x,y) \leftrightarrow right(y,x)$, and {\tt touching} is necessarily a reflexive relation.  QSR representations in VoxML rely on composition and closure to infer new spatial propositional content, either for planning or querying about actions; thus if $left(block1,block7)$ exists, $right(block7,block1)$ must be created axiomatically, and if a move then creates $right(block6,block7)$, $right(block6,block1)$ must also be created by transitive closure, as must $left(block1,block6)$.

\begin{table}
\centering
{\tiny
\begin{tabular}{|ll|}
    \hline
    right block7 block1 & right,touching block6 block7\\
	touching block3 block1 & right block5 block1\\
	left block1 block5 & under,touching,support block7 block5\\
	left block1 block7 & under,touching,support block1 block3\\
	under,touching,support block3 block4 & touching block5 block7\\
	touching block6 block5 & right block5 block3\\
	under block1 block4 & block7 $<$359.883; 1.222356; 359.0561$>$\\
	touching block4 block3 & block1 $<$0; 0; 0$>$\\
	left block3 block5 & block6 $<$0.1283798; 359.5548; 0.9346825$>$\\
	left block1 block6 & block3 $<$0; 0; 0$>$\\
	left,touching block7 block6 & block5 $<$0; 0; -2.970282E-08$>$\\
	right block6 block1 & block4 $<$0; 0; 0$>$\\
    \hline
    \end{tabular}}
\caption{\label{table:db}Example relation set.  Blocks are labeled by VoxSim-internal names.  As blocks may have been turned during construction, we also stored rotations.}
\end{table}

In all 17 cases, at least one human judged the resulting structure to be an acceptable staircase.  However, the configuration and relative placement of the blocks at large vary across the dataset and the individual structures themselves are not all isomorphic to each other, making this both sparse and noisy data to train on.  Can an algorithm, then, infer the commonalities across this relatively small number of structures, and reliably reproduce them?

\section{Learning Framework}
\label{sec:learning}

We adopt deep learning methods to effectively parse the complexity of a continuous quantitative search space for goal selection, and the compactness of qualitative spatial representation to assess potential moves using heuristics.

Using the Keras framework \cite{chollet2015keras} with the TensorFlow backend \cite{abadi2016tensorflow}, we trained models on the relations extracted from the examples to generate moves that would create a structure intended to approximate the training data. Inferences we hoped the learning method would be able to make include: 1) Individual blocks are interchangeable in the overall structure; 2) Overall orientation of the structure is arbitrary; 3) Progressively higher stacks of blocks in one direction are required.

We enforced constraints that {\it each block may only be moved once} and {\it once a block is placed in a relation, that relation may not be broken}.

\paragraph{First move selection.}

The first move made in the course of building such a structure may effectively be random, as blocks are intended to be interchangeable and a variety of relations might be created between them to form part of the goal structure.  However, in order to sample from the actual training data (and to avoid artificially introducing inference number 1 above to the final model), we trained a multi-layer perceptron (MLP) with 4 hidden dense layers of 64 nodes and ReLU activation, RMSProp optimization and sigmoid activation on the output layer that, given a random choice of two distinct blocks, returns the most likely relation between them.  Input is a pair of indices ranging from 0-5 denoting the blocks.  Output is the index of the relation to be created between them, out of all relations observed in the training data (12 total from this dataset).  This forms the first move: a string of the form $put(blockX,rel(blockY))$ added to the list of moves to be executed when the algorithm is complete.

\paragraph{Reference example selection.}

With each successive move, the system updates the relations in force between all placed blocks, and must predict an example of a completed structure it believes the generated moves are approaching.

Since the relations are stored as an unordered list and may be recreated in effectively any order during the new structure generation phase, we predict the reference example at each step using a convolutional neural network (CNN) with 4 1D convolution layers (64 nodes and ReLU activation on layers 1 and 2, 128 nodes and ReLU activation on layers 3 and 4, and RMSProp optimization), 1D max pooling after the 2nd and 4th layers, and a 50\% dropout layer before the output (softmax) layer.  CNNs have demonstrated utility in a variety of image recognition and natural language processing tasks (cf. \citeauthor{xu2014deep} \citeyear{xu2014deep}; \citeauthor{hu2014convolutional} \citeyear{hu2014convolutional}), and as this step of the learning algorithm is intended to predict a spatial configuration from a sequential representation, a CNN is an apt choice for this task.  Input is a sequence of triples ($x$,$y$,$r$) where $x$ and $y$ are indices of blocks and $r$ is the index of the relation between them (e.g., for two adjacent blocks, the input might look like $[(0,1,2),(1,0,5)]$ where 2 and 5 are the indices for $left$ and $right$, respectively).  Output is the index of the predicted example in the database, which is then retrieved as a set of similarly formatted block-block-relation triples.  The CNN is unlikely to be a very accurate predictor of final configuration after the first move(s) but as the search space shrinks from subsequent moves, it should become more so.

This predicted example serves as the goal state for the heuristic step.  It may change after each move as the currently-generated configuration more closely resembles different examples from those chosen previously.  At each step, the example's relation set is stored to compare to the next move prediction.

\paragraph{Next move prediction.}

Having chosen a reference example at the current step, we must then predict what next moves can be made that would bring us closer to that example.  For this step, we use a 3-layer long short-term memory (LSTM) network with 32 nodes in each layer and RMSProp optimization, with softmax activation over $n$ timesteps, where $n$ is the longest number of relations stored for any single example (over this data, $n$=20).  LSTMs have demonstrated utility in natural language processing and event recognition tasks due to their ability to capture long distance dependencies in sequences \cite{hochreiter1997long}.

We train the LSTM on ``windows" of block-block-relation sets taken from the training data, formatted the same as the CNN inputs, which range in length from $1$ to $max-1$ where $max$ is the length of the relation set for a given example.  Window length increases with the timestep.  These relation sets are indexed and trained relative to the ``holdout," or complementary relation set, also of length in the range $1$ to $max-1$ (decreasing as the timestep increases).  Since the relations are stored in an unordered manner, we need to capture all combinations in a given example, which makes the LSTM the most time-intensive portion of the model training, although given the small size of the data ($\sim$2000 configuration-holdout pairs) this is not unreasonable.\footnote{Training was conducted without GPUs, and the decrease in categorical cross-entropy loss appears to taper off after 20 epochs, or about a minute of training on the hardware used.}

Since not all combinations of blocks and relations occur in the training data, the input to the LSTM in the prediction step is a heuristically-determined ``closest match" to the configuration resulting from the previous moves.  This is the first of two uses of heuristics to prune the search space.  The LSTM predicts a ``holdout" or remaining set of block-block relations to create to approach the CNN-selected example, given the current configuration.  Heuristics are assessed for which selects the best moves toward the CNN-chosen goal state from these presented move options.

\subsection{Heuristic Estimation and Pruning}

With the selection of an example and a predicted set of next moves, we can focus only on the relations that we wish to create in the next step, independent of the blocks involved.  First we take the intersection of the relations that occur in the example and the holdout set (allowing for repetitions, such that if a relation occurs twice in both the example and the holdout set, it occurs twice in the intersection).  Then, for each relation in this intersection set that can still be created in the current configuration (i.e., a block can only have another block placed on it if its top surface is not covered), we try adding it to the current set of relations and calculating the heuristic distance between that combination of relations (with closure) and the relations in the example case.  We examined 5 different heuristics: random chance, two standard distance metrics, a custom graph-matching algorithm detailed subsequently, and a combined method.  All except random chance use the output of the learning pipeline to define the search space topology.
\begin{enumerate}
\item {\it Chance}.  As a baseline, we disregarded the neural networks' output and chose the next move at each step by random chance.  A block to place was selected randomly, and then a relation to create was also selected randomly out of the set of available placements on the current structure.  This allowed us to assess the odds of the agent approaching a staircase structure by random guessing.
\item {\it Jaccard distance (JD)}.  Jaccard index ($J$) \cite{jaccard1912distribution} is a well-known metric to compare sets of tokens.  Since we are interested in the {\it lowest} heuristic distance between two sets, we use the Jaccard {\it distance} or $1-J$.
\item {\it Levenshtein distance (LD)}.  Since Jaccard distance only accounts for the presence or absence of a token, and we want to preserve the full set of relations (i.e., the same block may have two blocks under it, or to the left of it), we use the Levenshtein distance (LD) \cite{levenshtein1966binary}.  Since Levenshtein distance leaves the arbitrary order of the tokens intact, we sort the list alphabetically first to fix the same relative order in all cases.
\item {\it Graph matching}.  Simpler heuristics run a high risk of choosing an inappropriate relation outright due to low distance with a final result even though creating some substructure that uses a different additional relation may be the best move on the way to the final structure.  Graph matching alleviates this by taking into account common subgraphs.  The graph matching algorithm we use is discussed in more detail below.
\item {\it LD-pruned graph matching}.  In principle, since graph matching allows for a choice out of all possible moves generated by the LSTM, it may occasionally risk choosing an inappropriate move since the relations may be drawn from a larger set (i.e., the actual best move creates an $on$ relation, but the options given may also contain a number of $left$ and $right$ relations, one of which may be chosen by graph matching alone), we also tried pruning the set of move options presented to the graph matcher to ones containing the ``best" destination block for each relation option and then choosing the move with graph matching.
\end{enumerate}

Using the steps outlined above, moves are generated until all blocks have been placed, resulting in a sequence like the following:\\
\noindent$put(block6,left(block4));put(block5,rightdc(block4));$
\noindent
$put(block7,on(block4));put(block1,on(block6));$
\noindent
$put(block3,on(block1))$

Here, $left$ and $right$ are operationalized as ``left and touching" and ``right and touching," respectively, while $leftdc$ and $rightdc$ are interpreted as left and right and $DC$ (``disconnected" in the Region Connection Calculus).

\subsection{Action Selection with Heuristic Graph-Matching}
\label{ssec:graph-matching}

After the learning pipeline has generated possible actions, the graph matcher selects one as follows:
\begin{enumerate}
\item For each potential action, compute a distinct \emph{state graph} of qualitative spatial relations that would hold \emph{if the action succeeded}.
\item Compute the \emph{maximal common subgraph} (MCS) of each state graph against a qualitative spatial relation graph of the agent's goal (i.e., the structure it intends to build).
\item Choose the action with the highest-scoring MCS with the agent's goal.
\end{enumerate}
We use an implementation written on the \spire reasoning system \cite{r3_acs_2017} to perform these operations.
\spire includes a relational knowledge base to store the contents of the agent's goals and potential actions, and it implements a \emph{structure-mapping} algorithm \cite{forbus2017extending} to compute MCSs over these relational graphs.
Structure-mapping has been used for agent action-selection \cite{friedman2017analogical} as well as large-scale cognitive modeling \cite{forbus2017extending} and learning spatial structures and anatomical concepts via generalization and comparative analysis \cite{mclure2015extending}.

\begin{figure}[h!]
\centering
\includegraphics[width=0.45\textwidth]{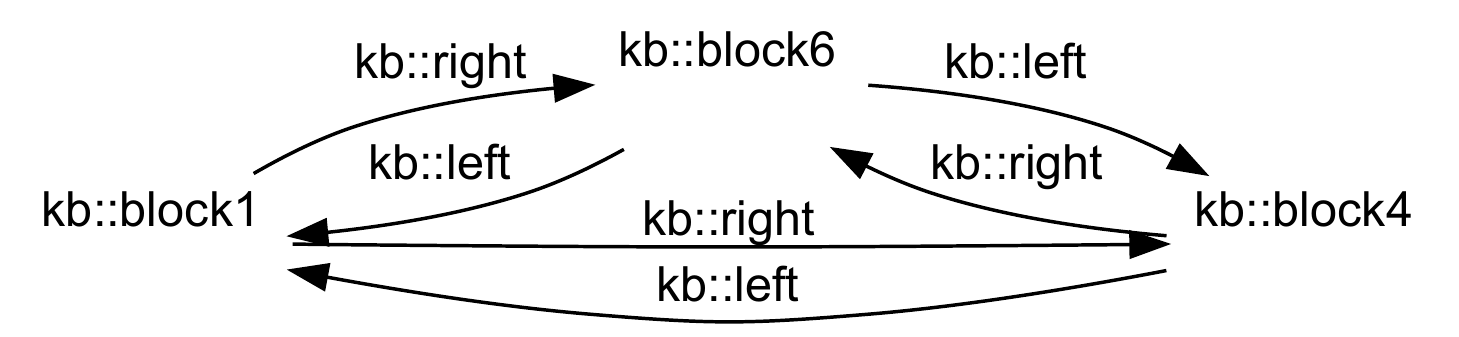}
\caption{Graph representation of possible action result}
\label{fig:potential-graph}
\end{figure}

\begin{figure}[h!]
\centering
\includegraphics[width=0.45\textwidth]{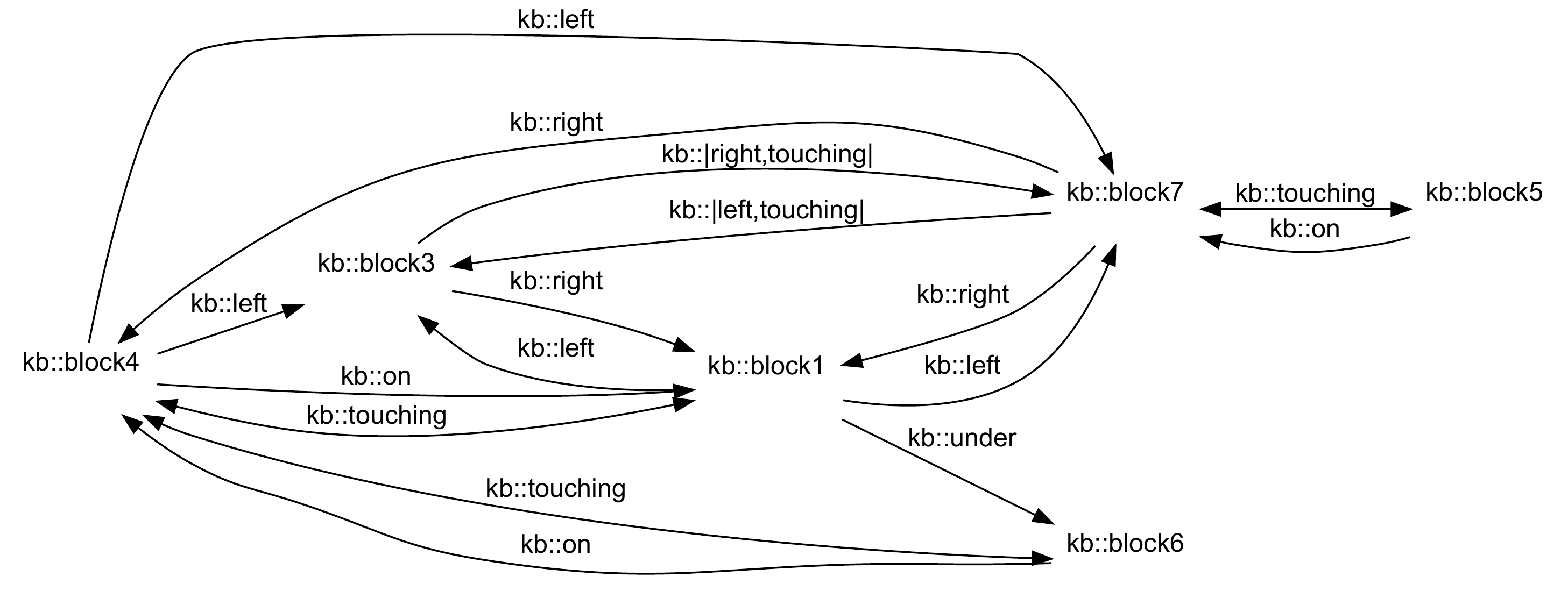}
\caption{Graph representation of goal configuration}
\label{fig:target-graph}
\end{figure}

Given two relational graphs---for instance, a possible action result (Fig.~\ref{fig:potential-graph}) and the goal configuration (Fig.~\ref{fig:target-graph})---the structure-mapping algorithm first scores \emph{local} solutions by the number of consistent relations supported by each node-to-node (i.e., block-to-block) correspondence across graphs.
It then uses a greedy algorithm over these local solutions to accrue a global subgraph isomorphism that approximates a MCS over the two graphs.

Each \spire MCS includes a numerical \emph{similarity score} describing the size of the MCS, which we use to approximate graph similarity.  Importantly, structure-mapping is not influenced by symbols or nodes themselves (e.g., block IDs), so blocks are perfectly interchangeable in this graph-matching phase.  The vocabulary of the graph-matcher is not specified \emph{a priori}, so the relational vocabulary can be elaborated with novel relations and the graph-matcher will match new relations according to its structure-mapping algorithm.

The graph matcher selects the action with the maximum-scoring MCS against the goal, biasing it to select actions that produce more shared qualitative spatial relations with the goal state, all else being equal.  The graph-matcher uses no predefined operators, axioms, or plans in this action selection phase.

\section{Evaluation and Results}

The generated move sequence can then be input into VoxSim again to command the virtual agent to actually construct the structure generated from the trained model.

Using the VoxSim system, we generated 50 structures using this structure learning and generation system---10 structures using each heuristic method, including random chance as a baseline---and presented them to 8 annotators for evaluation.  All annotators are adult English speakers with a college degree.  They were given pictures of each structure and asked to answer the question: {\it ``On a scale of 0-10 (10 being best), how much does the structure shown resemble a staircase?"}  No extra information was given, specifically in the hopes that the annotator would answer based on their particular notion of a canonical or prototypical staircase.  The order in which each evaluator viewed the images was randomized, in order to lessen the overall possible effect of evaluator judgments on the early examples affecting their judgments on the later ones.  In Table~\ref{table:eval}, {\it Chance} refers to the baseline, {\it JD} to the Jaccard distance heuristic, {\it LD} to the Levenshtein distance heuristic, {\it \spire} to the graph matching system, and {\it Comb.} to the combination of LD and \spire.

\begin{table}[h!]
\centering
\begin{tabular}{|l|l|l|}
\hline
{\bf Heuristic} & {\bf Avg. Score} ($\mu$) & {\bf Std. Dev.} ($\sigma$) \\
\hline
Chance & 2.0375	& 1.0122 \\
\hline
JD & 4.3375	& 2.0387 \\
\hline
LD & 3.7688 & 2.1028 \\
\hline
\spire & 5.8313 & 2.7173 \\
\hline
Comb. & 4.7188 & 2.4309 \\
\hline
\end{tabular}
\caption{\label{table:eval}Evaluator judgments of generated staircase quality by heuristic algorithm}
\end{table}
The average score ($\mu$) is given for all evaluations over all structures generated using a given heuristic and can be used to assess the quality of structures generated using that heuristic (or alternately, the probability of that heuristic generating quality 3-step staircases when given example structures and potential moves).  The standard deviation ($\sigma$) is the standard deviation of the average scores of each structure generated using the given heuristic.  This can be considered an overall representation of how certain evaluators were of the quality of structures generated using that heuristic (lower $\sigma$ corresponds to greater overall evaluator agreement).

\begin{figure}[h!]
\centering
\begin{tabular}{|l|l|l|}
\hline
Chance: & 
\includegraphics[height=.5in]{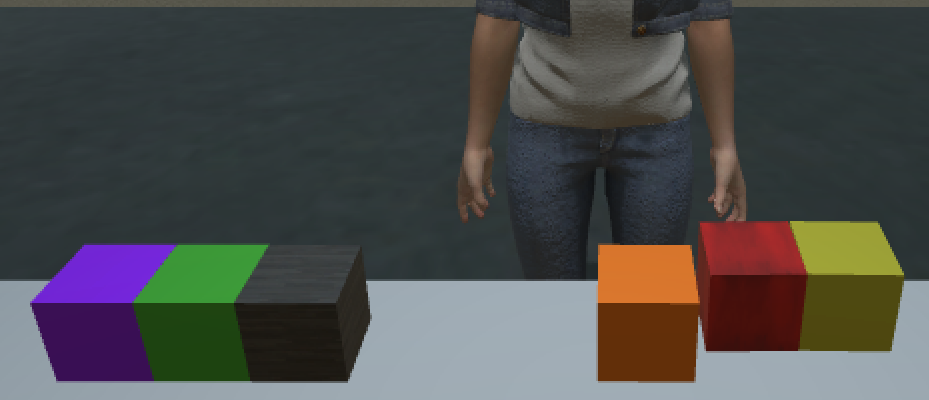} &
\includegraphics[height=.5in]{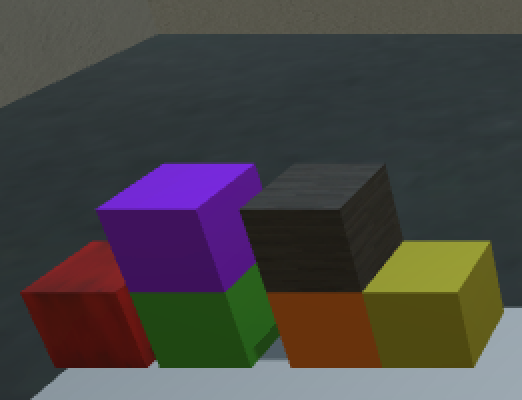} \\
JD: & 
\includegraphics[height=.5in]{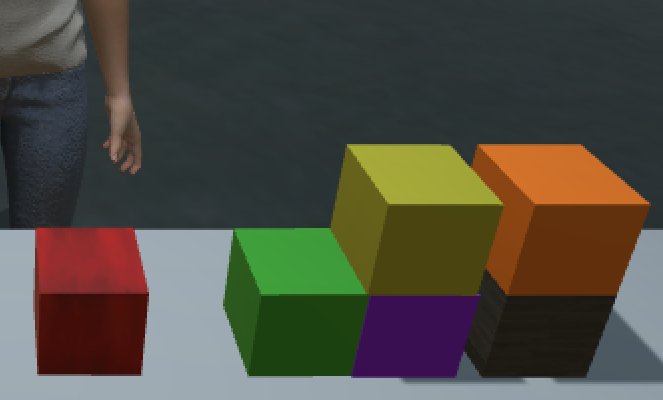} &
\includegraphics[height=.5in]{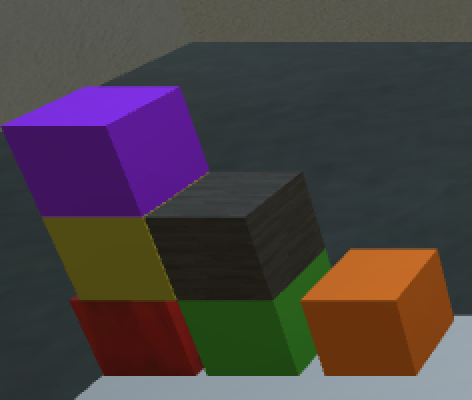} \\
LD: &
\includegraphics[height=.5in]{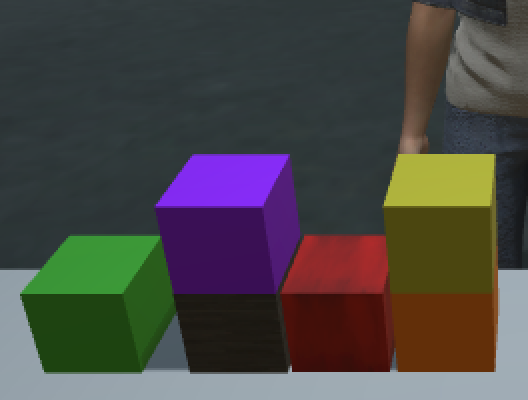} &
\includegraphics[height=.5in]{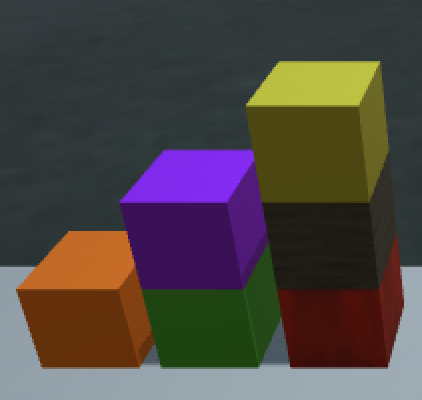} \\
\spire: &
\includegraphics[height=.5in]{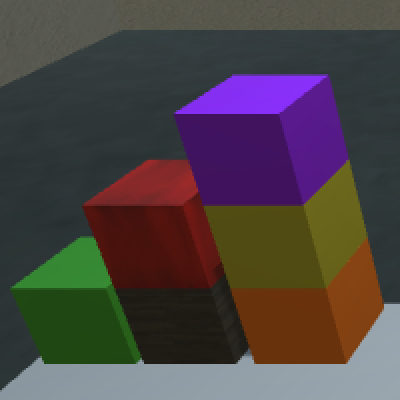} &
\includegraphics[height=.5in]{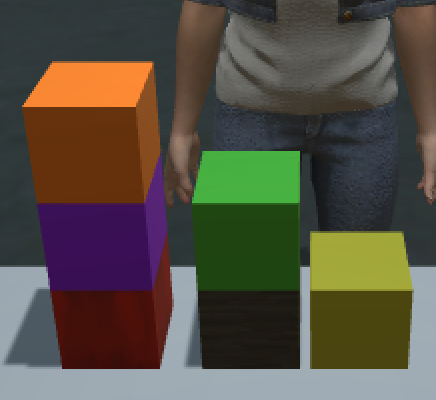} \\
Comb.: &
\includegraphics[height=.5in]{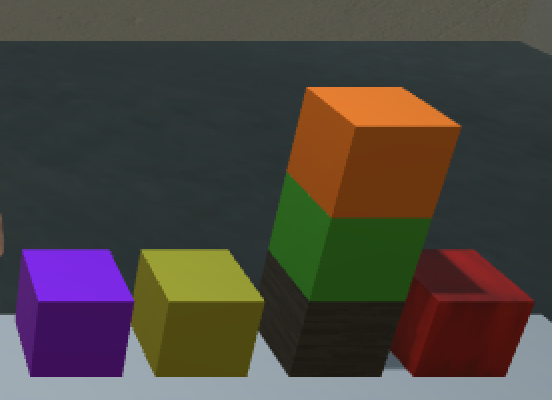} &
\includegraphics[height=.5in]{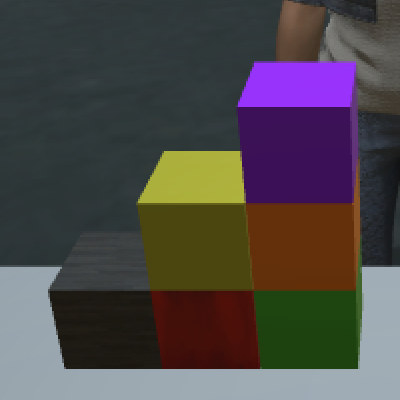} \\
\hline
\end{tabular}
\caption{Median- (L) and highest-scored (R) structure generated using each heuristic (average evaluator score)}
\label{fig:results}
\end{figure}

Using \spire for heuristic graph matching generated the most highly-rated structures on average, followed by the combined method, Jaccard distance, and Levenshtein distance.  All methods improved significantly on the baseline.

Pruning the presented move options by Levenshtein distance before applying graph matching performed better than LD alone but did not approach the performance of graph matching alone.  This suggests that constraining the search space too much sometimes unduly throws out the best move options.  Given that Levenshtein distance underperformed Jaccard distance, more research is needed to see if combining Jaccard distance with graph matching could improve results further, or would simply outperform the JD heuristic while still underperforming pure graph matching.

\begin{figure}[h!]
\centering
\includegraphics[height=1.85in]{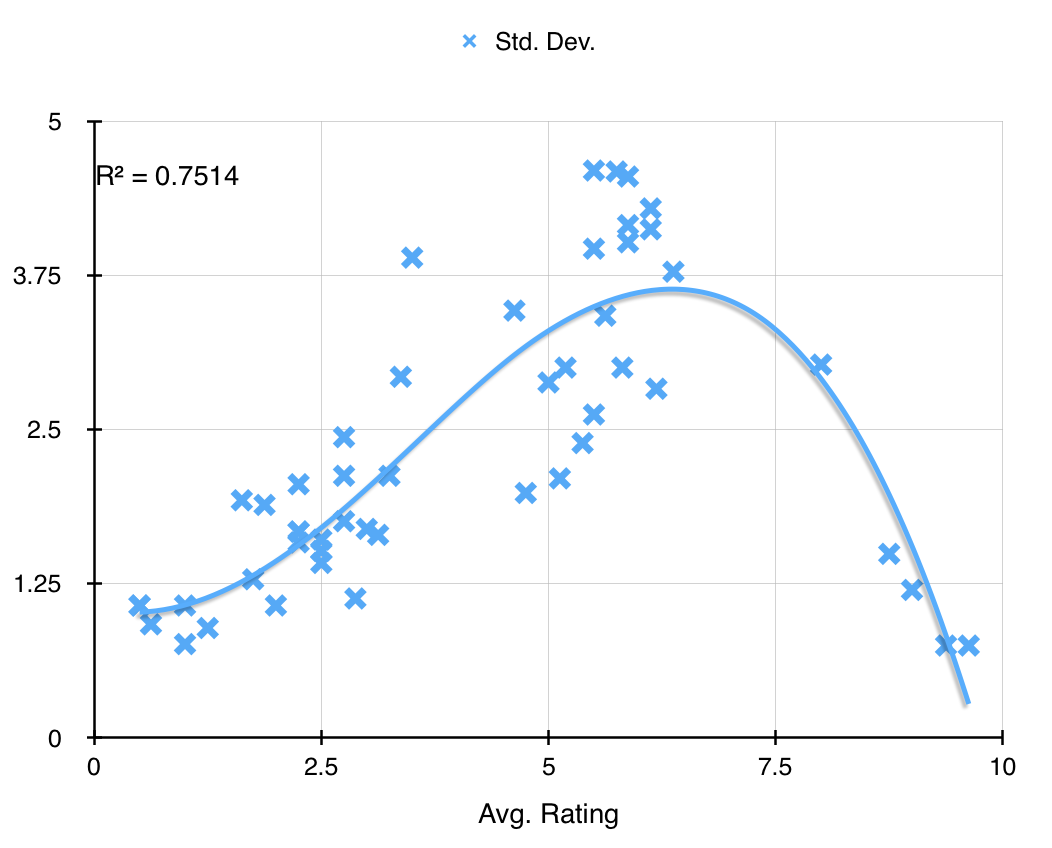}
\caption{Average rating vs. rating standard deviation of each generated structure}
\label{fig:avg-vs-std}
\end{figure}

Fig.~\ref{fig:avg-vs-std} plots $\sigma$ against $\mu$ for each structure, with a cubic best-fit line chosen by least-squares deviances (R$^2$ = 0.7514), and we can see that evaluators tended to agree most on very well-constructed staircases, and more on obvious ``non-staircases'' than on the middle cases.  For very low- or very high-scored examples, $\sigma$ is much lower---often near zero---than for mid-scored examples, suggesting stronger annotator agreement, even with only 8 judges, on ``good'' staircases vs. middle cases that displayed some but not all inferences desired in Learning Framework.

\subsection{Discussion}
Previously we outlined some inferences we hoped the learning pipeline would be able to make from the sample data.  It appears in novel generated examples, that {\it individual blocks are interchangeable}, as many examples (both correct and incorrect) consisted of identical configurations constructed of differently-colored blocks.  The system appeared successful in producing structures of {\it arbitrary orientation}, that is, both left- and right-pointing staircases.  {\it left} and {\it right} were the only directional relations available in the training data.  Graph matching appeared to be most successful at pruning the options down to moves that would result in {\it progressively higher stacks of blocks in a single direction}.  Sometimes the system generated a ``near-staircase'' structure of columns in a 1-3-2 configuration instead of the desired 1-2-3, and other times built ``staircases'' of two levels (1 block high and 2 blocks high, or 2 blocks high and 4 blocks high).

\begin{figure}[h!]
\centering
\includegraphics[height=.6in]{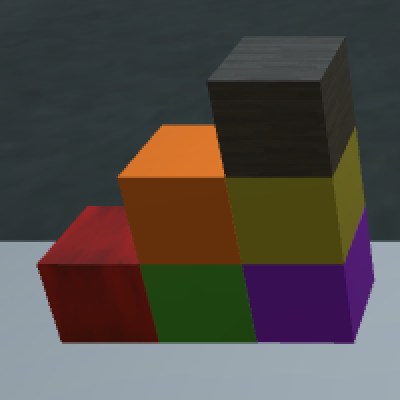}
\includegraphics[height=.6in]{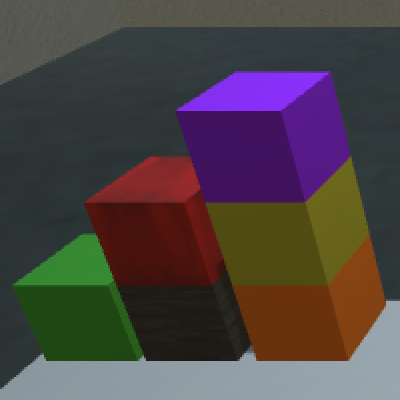}
\includegraphics[height=.6in]{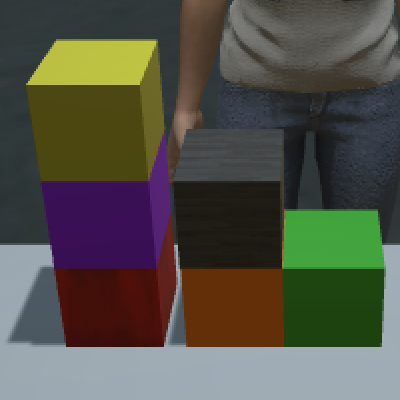}
\includegraphics[height=.6in]{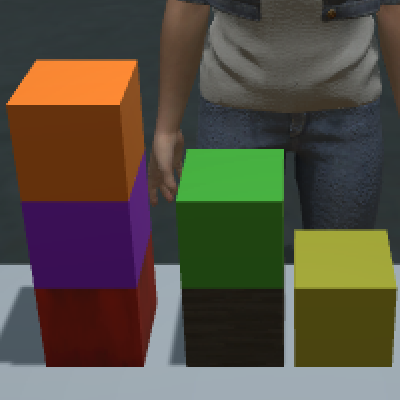}
\caption{Generated staircases displaying desired inferences}
\end{figure}

\section{Improving the Model}
\label{sec:improving}

Even with the highest-performing heuristic, the system sometimes generates structures that are judged incorrect by evaluators.  Some of these are due to downstream errors from the CNN's predicted configuration which, when intersected with the LSTM's predicted holdout configuration, does not produce any possible moves that approach a 3-step staircase, but the algorithm is required to choose one anyway.  Examples include putting a third block on the center (2-step) column, or creating a 4-block row on the bottom level where there should be a maximum of 3 blocks.

Since we do not currently check for this condition, it is possible that the learning agent is generating moves that it views as optimal in the short term but prove counterproductive in the long term.  Better long-term moves may reside in the intersection of lower-ranked CNN and LSTM choices, and we could improve this by looking in the $n$-best results from the neural nets rather than just top-ranked results.

In addition, some of the constraints we impose on the system are counterproductive in the long term but do not allow for correction.  For instance, once a block has been placed, there is a hard-coded constraint prohibiting it from being moved, meaning that any subsequent move must involve placing a new block instead of moving a previously-placed one to a better position, even when that would more optimally complete the desired structure.  Allowing for backtracking and re-planning would lessen these problems.

\section{Future Work}

The trained model can be stored in a database for subsequent retrieval under a label of choice.  When the agent is again asked to ``build a staircase,'' it can retrieve the model from the database and, along with graph matching, have a ready blueprint to build an example of what it understands to be a staircase based on the ones it has observed.

Given the interactive nature of the task used to gather the example structures, we envision additions to the interaction and learning framework that would allow turning a negative example into a corresponding positive example, which increases the overall data size and introduces a minimal distinction between a good example and a bad one.  The still-small sample size facilitates online learning.

Given a generated structure as in the bottom left of Fig.~\ref{fig:results}, where if the red block were moved on top of the yellow block, the overall structure would better satisfy the constraints of a staircase, an example correction interaction might proceed as follows, depending on what input modalities the interaction system handles:

\begin{dialogue}
\speak{Avatar} {\bf Is this a staircase?}
\speak{Human} No. \direct{Current configuration stored as negative example.}  Pick up the red block. \direct{{\sc human} points at the red block.}
\speak{Avatar} Okay.  \direct{{\sc avatar} picks up the red block.}
\speak{Human} Put it on the yellow block. \direct{{\sc human} points at the yellow block.}
\speak{Avatar} Okay.  \direct{{\sc avatar} puts the red block on the yellow block}
\speak{Human} {\bf This is a staircase}. \direct{New structure stored as positive example contrasting to previous structure.} 
\end{dialogue}

Such a format would allow us to examine how {\it small} a sample set this method requires. A model trained on one example would likely replicate that same example. Generalizing over the desired inferences would require instruction in the form of a dialogue. Models could be transferred to a new label and then corrected/updated by instruction.

Our method depends on three components: a convolutional neural net to predict an example target configuration after each move, a long short-term memory network to predict the remaining moves needed to get there, and an appropriate heuristic function to choose the best moves out of the options presented that can legally be made in the present situation.  It has the advantage of functioning on a small sample size and so does not require much in training time or resources, and appears to be fairly successful in generating structures that comport with a human's notion of the intended label, but also leaves some questions unanswered:

Can this method generalize to other shapes (e.g., a pyramid), or more complex configurations, given the proper relational primitives?  In the sample data, all the user-generated examples consisted of staircases oriented along the left-right (X) axis, even though the the concept of a staircase could also be aligned along the back-to-front (Z) axis.  How would the addition of {\tt in\_front} and {\tt behind} to the vocabulary expand the search space, and what other methods would be needed to ensure quality?

Can we use this to generalize further over an introduced concept, particularly in situations where the domain space provides room for the search space to expand beyond that given in the examples \cite{leake2015flexible}?  Since, even in many examples given low scores by the evaluators (e.g., Fig.~\ref{fig:results}, 3rd row left), the system appeared to generate some sort of stepped structure, perhaps this concept can successfully be generalized, and so if ten blocks were on the table and the system told to proceed until all blocks were placed, would it be able to create a 4-step staircase out of ten blocks?

\section{Conclusion}

This paper discusses a procedure for observing sparse and noisy examples of a structure previously unknown to an AI agent, and using them to generate new examples of structures that share the same qualities.  We leverage the strengths of deep learning to select examples out of noisy data and use heuristic functions to prune the resulting search space.  Fusing qualitative representations with deep learning requires significantly less overhead in terms of data and training time than many traditional machine learning approaches.

Deep learning, of course, is just one method of learning constraints and there are others, such as inductive logic programming, which could be equally effective at managing the search space and are worth examination in conjunction with qualitative representations.  We have examined different heuristics in the generation phase and discovered that graph matching tends to yield the best results according to human evaluators.  That qualitative representation seems to be effective in this procedural problem-solving task supports other evidence that qualitative spatial relations are also effective in recognition or classification tasks, as indicated by \citeauthor{hawes2012towards} (\citeyear{hawes2012towards}) and \citeauthor{kunze2014combining} (\citeyear{kunze2014combining}), among others, which is critical to completely teaching a new structural concept to an AI agent.

In human-computer interactions, novel concepts should be able to be introduced in real time, and we believe the structure learning method described here can be deployed in a human-computer interaction to create new positive examples and correct negative ones, allowing for integration of online and reinforcement learning.  Since humans typically assess spatial relations qualitatively, not quantitatively \cite{davis2016scope}, any AI aspiring to human-like domains should also perform well on qualitative data; this paper provides further empirical evidence in favor of this.

\bibliography{Krishnaswamy-MasterReferences}

\begin{thebibliography}{}

\bibitem[\protect\citeauthoryear{Abadi \bgroup et al\mbox.\egroup
  }{2016}]{abadi2016tensorflow}
Abadi, M.; Barham, P.; Chen, J.; Chen, Z.; Davis, A.; Dean, J.; Devin, M.;
  Ghemawat, S.; Irving, G.; Isard, M.; et~al.
\newblock 2016.
\newblock Tensorflow: A system for large-scale machine learning.
\newblock In {\em Proceedings of the 12th USENIX Symposium on Operating Systems
  Design and Implementation (OSDI). Savannah, Georgia, USA}.

\bibitem[\protect\citeauthoryear{Alayrac \bgroup et al\mbox.\egroup
  }{2016}]{alayrac2016unsupervised}
Alayrac, J.-B.; Bojanowski, P.; Agrawal, N.; Sivic, J.; Laptev, I.; and
  Lacoste-Julien, S.
\newblock 2016.
\newblock Unsupervised learning from narrated instruction videos.
\newblock In {\em Proceedings of the IEEE Conference on Computer Vision and
  Pattern Recognition},  4575--4583.

\bibitem[\protect\citeauthoryear{Albath \bgroup et al\mbox.\egroup
  }{2010}]{albath2010rcc}
Albath, J.; Leopold, J.~L.; Sabharwal, C.~L.; and Maglia, A.~M.
\newblock 2010.
\newblock R{CC-3D}: Qualitative spatial reasoning in 3{D}.
\newblock In {\em CAINE},  74--79.

\bibitem[\protect\citeauthoryear{Alomari \bgroup et al\mbox.\egroup
  }{2017a}]{alomari2017learning}
Alomari, M.; Duckworth, P.; Hogg, D.~C.; and Cohn, A.~G.
\newblock 2017a.
\newblock Learning of object properties, spatial relations, and actions for
  embodied agents from language and vision.
\newblock In {\em The AAAI 2017 Spring Symposium on Interactive Multisensory
  Object Perception for Embodied Agents Technical Report SS-17-05},  444--448.
\newblock AAAI Press.

\bibitem[\protect\citeauthoryear{Alomari \bgroup et al\mbox.\egroup
  }{2017b}]{al2017natural}
Alomari, M.; Duckworth, P.; Hogg, D.~C.; and Cohn, A.~G.
\newblock 2017b.
\newblock Natural language acquisition and grounding for embodied robotic
  systems.
\newblock In {\em AAAI},  4349--4356.

\bibitem[\protect\citeauthoryear{Asada, Uchibe, and
  Hosoda}{1999}]{asada1999cooperative}
Asada, M.; Uchibe, E.; and Hosoda, K.
\newblock 1999.
\newblock Cooperative behavior acquisition for mobile robots in dynamically
  changing real worlds via vision-based reinforcement learning and development.
\newblock {\em Artificial Intelligence} 110(2):275--292.

\bibitem[\protect\citeauthoryear{Barbu \bgroup et al\mbox.\egroup
  }{2012}]{barbu2012simultaneous}
Barbu, A.; Michaux, A.; Narayanaswamy, S.; and Siskind, J.~M.
\newblock 2012.
\newblock Simultaneous object detection, tracking, and event recognition.
\newblock {\em arXiv preprint arXiv:1204.2741}.

\bibitem[\protect\citeauthoryear{Binong and
  Hazarika}{2018}]{binong2018extracting}
Binong, J., and Hazarika, S.~M.
\newblock 2018.
\newblock Extracting qualitative spatiotemporal relations for objects in a
  video.
\newblock In {\em Proceedings of the International Conference on Computing and
  Communication Systems},  327--335.
\newblock Springer.

\bibitem[\protect\citeauthoryear{Chollet}{2015}]{chollet2015keras}
Chollet, Fran{\c{c}}ois, e.~a.
\newblock 2015.
\newblock Keras.

\bibitem[\protect\citeauthoryear{Craw, Wiratunga, and
  Rowe}{2006}]{craw2006learning}
Craw, S.; Wiratunga, N.; and Rowe, R.~C.
\newblock 2006.
\newblock Learning adaptation knowledge to improve case-based reasoning.
\newblock {\em Artificial Intelligence} 170(16-17):1175--1192.

\bibitem[\protect\citeauthoryear{Das \bgroup et al\mbox.\egroup
  }{2017}]{das2017embodied}
Das, A.; Datta, S.; Gkioxari, G.; Lee, S.; Parikh, D.; and Batra, D.
\newblock 2017.
\newblock Embodied question answering.
\newblock {\em arXiv preprint arXiv:1711.11543}.

\bibitem[\protect\citeauthoryear{Davis and Marcus}{2016}]{davis2016scope}
Davis, E., and Marcus, G.
\newblock 2016.
\newblock The scope and limits of simulation in automated reasoning.
\newblock {\em Artificial Intelligence} 233:60--72.

\bibitem[\protect\citeauthoryear{Dubba \bgroup et al\mbox.\egroup
  }{2015}]{dubba2015learning}
Dubba, K.~S.; Cohn, A.~G.; Hogg, D.~C.; Bhatt, M.; and Dylla, F.
\newblock 2015.
\newblock Learning relational event models from video.
\newblock {\em Journal of Artificial Intelligence Research} 53:41--90.

\bibitem[\protect\citeauthoryear{Fernando, Shirazi, and
  Gould}{2017}]{fernando2017unsupervised}
Fernando, B.; Shirazi, S.; and Gould, S.
\newblock 2017.
\newblock Unsupervised human action detection by action matching.
\newblock In {\em Proceedings of the IEEE Conference on Computer Vision and
  Pattern Recognition Workshops},  1--9.

\bibitem[\protect\citeauthoryear{Forbus \bgroup et al\mbox.\egroup
  }{2017}]{forbus2017extending}
Forbus, K.~D.; Ferguson, R.~W.; Lovett, A.; and Gentner, D.
\newblock 2017.
\newblock Extending sme to handle large-scale cognitive modeling.
\newblock {\em Cognitive Science} 41(5):1152--1201.

\bibitem[\protect\citeauthoryear{Friedman \bgroup et al\mbox.\egroup
  }{2017a}]{r3_acs_2017}
Friedman, S.; Burstein, M.; McDonald, D.; Plotnick, A.; Bobrow, L.; Bobrow, R.;
  Cochran, B.; and Pustejovsky, J.
\newblock 2017a.
\newblock Learning by reading: Extending and localizing against a model.
\newblock {\em Advances in Cognitive Systems} 5:77--96.

\bibitem[\protect\citeauthoryear{Friedman \bgroup et al\mbox.\egroup
  }{2017b}]{friedman2017analogical}
Friedman, S.; Burstein, M.; Rye, J.; and Kuter, U.
\newblock 2017b.
\newblock Analogical localization: Flexible plan execution in open worlds.
\newblock {\em ICCBR 2017 Computational Analogy Workshop}.

\bibitem[\protect\citeauthoryear{Gatsoulis \bgroup et al\mbox.\egroup
  }{2016}]{gatsoulis2016qsrlib}
Gatsoulis, Y.; Alomari, M.; Burbridge, C.; Dondrup, C.; Duckworth, P.;
  Lightbody, P.; Hanheide, M.; Hawes, N.; Hogg, D.; Cohn, A.; et~al.
\newblock 2016.
\newblock Qsrlib: a software library for online acquisition of qualitative
  spatial relations from video.

\bibitem[\protect\citeauthoryear{Gergely, Bekkering, and
  Kir{\'a}ly}{2002}]{gergely2002developmental}
Gergely, G.; Bekkering, H.; and Kir{\'a}ly, I.
\newblock 2002.
\newblock Developmental psychology: Rational imitation in preverbal infants.
\newblock {\em Nature} 415(6873):755.

\bibitem[\protect\citeauthoryear{Gullapalli}{1990}]{gullapalli1990stochastic}
Gullapalli, V.
\newblock 1990.
\newblock A stochastic reinforcement learning algorithm for learning
  real-valued functions.
\newblock {\em Neural networks} 3(6):671--692.

\bibitem[\protect\citeauthoryear{Hart, Nilsson, and
  Raphael}{1968}]{hart1968formal}
Hart, P.~E.; Nilsson, N.~J.; and Raphael, B.
\newblock 1968.
\newblock A formal basis for the heuristic determination of minimum cost paths.
\newblock {\em IEEE transactions on Systems Science and Cybernetics}
  4(2):100--107.

\bibitem[\protect\citeauthoryear{Hawes \bgroup et al\mbox.\egroup
  }{2012}]{hawes2012towards}
Hawes, N.; Klenk, M.; Lockwood, K.; Horn, G.~S.; and Kelleher, J.~D.
\newblock 2012.
\newblock Towards a cognitive system that can recognize spatial regions based
  on context.
\newblock In {\em AAAI}.

\bibitem[\protect\citeauthoryear{Hermann \bgroup et al\mbox.\egroup
  }{2017}]{hermann2017grounded}
Hermann, K.~M.; Hill, F.; Green, S.; Wang, F.; Faulkner, R.; Soyer, H.;
  Szepesvari, D.; Czarnecki, W.; Jaderberg, M.; Teplyashin, D.; et~al.
\newblock 2017.
\newblock Grounded language learning in a simulated 3d world.
\newblock {\em arXiv preprint arXiv:1706.06551}.

\bibitem[\protect\citeauthoryear{Hochreiter and
  Schmidhuber}{1997}]{hochreiter1997long}
Hochreiter, S., and Schmidhuber, J.
\newblock 1997.
\newblock Long short-term memory.
\newblock {\em Neural computation} 9(8):1735--1780.

\bibitem[\protect\citeauthoryear{Hu \bgroup et al\mbox.\egroup
  }{2014}]{hu2014convolutional}
Hu, B.; Lu, Z.; Li, H.; and Chen, Q.
\newblock 2014.
\newblock Convolutional neural network architectures for matching natural
  language sentences.
\newblock In {\em Advances in neural information processing systems},
  2042--2050.

\bibitem[\protect\citeauthoryear{Jaccard}{1912}]{jaccard1912distribution}
Jaccard, P.
\newblock 1912.
\newblock The distribution of the flora in the alpine zone.
\newblock {\em New phytologist} 11(2):37--50.

\bibitem[\protect\citeauthoryear{Kordjamshidi \bgroup et al\mbox.\egroup
  }{2011}]{kordjamshidi2011relational}
Kordjamshidi, P.; Frasconi, P.; Van~Otterlo, M.; Moens, M.-F.; and De~Raedt, L.
\newblock 2011.
\newblock Relational learning for spatial relation extraction from natural
  language.
\newblock In {\em International Conference on Inductive Logic Programming},
  204--220.
\newblock Springer.

\bibitem[\protect\citeauthoryear{Krishnaswamy and
  Pustejovsky}{2016a}]{krishnaswamy2016multimodal}
Krishnaswamy, N., and Pustejovsky, J.
\newblock 2016a.
\newblock Multimodal semantic simulations of linguistically underspecified
  motion events.
\newblock In {\em Spatial Cognition X: International Conference on Spatial
  Cognition}.
\newblock Springer.

\bibitem[\protect\citeauthoryear{Krishnaswamy and
  Pustejovsky}{2016b}]{krishnaswamy2016voxsim}
Krishnaswamy, N., and Pustejovsky, J.
\newblock 2016b.
\newblock Vox{S}im: A visual platform for modeling motion language.
\newblock In {\em Proceedings of COLING 2016, the 26th International Conference
  on Computational Linguistics: Technical Papers}.
\newblock ACL.

\bibitem[\protect\citeauthoryear{Krishnaswamy and
  Pustejovsky}{2018}]{krishnaswamy2018LREC}
Krishnaswamy, N., and Pustejovsky, J.
\newblock 2018.
\newblock An evaluation framework for multimodal interaction.
\newblock {\em Proceedings of LREC}.

\bibitem[\protect\citeauthoryear{Kunze \bgroup et al\mbox.\egroup
  }{2014}]{kunze2014combining}
Kunze, L.; Burbridge, C.; Alberti, M.; Thippur, A.; Folkesson, J.; Jensfelt,
  P.; and Hawes, N.
\newblock 2014.
\newblock Combining top-down spatial reasoning and bottom-up object class
  recognition for scene understanding.
\newblock In {\em Intelligent Robots and Systems (IROS 2014), 2014 IEEE/RSJ
  International Conference on},  2910--2915.
\newblock IEEE.

\bibitem[\protect\citeauthoryear{Laird}{2012}]{laird2012soar}
Laird, J.~E.
\newblock 2012.
\newblock {\em The Soar cognitive architecture}.
\newblock MIT press.

\bibitem[\protect\citeauthoryear{Langley and Choi}{2006}]{langley2006unified}
Langley, P., and Choi, D.
\newblock 2006.
\newblock A unified cognitive architecture for physical agents.
\newblock In {\em Proceedings of the National Conference on Artificial
  Intelligence}, volume~21,  1469.
\newblock Menlo Park, CA; Cambridge, MA; London; AAAI Press; MIT Press; 1999.

\bibitem[\protect\citeauthoryear{Leake and Schack}{2015}]{leake2015flexible}
Leake, D., and Schack, B.
\newblock 2015.
\newblock Flexible feature deletion: compacting case bases by selectively
  compressing case contents.
\newblock In {\em International Conference on Case-Based Reasoning},  212--227.
\newblock Springer.

\bibitem[\protect\citeauthoryear{Levenshtein}{1966}]{levenshtein1966binary}
Levenshtein, V.~I.
\newblock 1966.
\newblock Binary codes capable of correcting deletions, insertions, and
  reversals.
\newblock In {\em Soviet physics doklady}, volume~10,  707--710.

\bibitem[\protect\citeauthoryear{Liang \bgroup et al\mbox.\egroup
  }{2018}]{liang2018visual}
Liang, K.; Guo, Y.; Chang, H.; and Chen, X.
\newblock 2018.
\newblock Visual relationship detection with deep structural ranking.

\bibitem[\protect\citeauthoryear{McLure, Friedman, and
  Forbus}{2015}]{mclure2015extending}
McLure, M.~D.; Friedman, S.~E.; and Forbus, K.~D.
\newblock 2015.
\newblock Extending analogical generalization with near-misses.
\newblock In {\em AAAI},  565--571.

\bibitem[\protect\citeauthoryear{M{\'e}nager}{2016}]{menager2016episodic}
M{\'e}nager, D.
\newblock 2016.
\newblock Episodic memory in a cognitive model.
\newblock In {\em ICCBR Workshops},  267--271.

\bibitem[\protect\citeauthoryear{Moratz, Nebel, and
  Freksa}{2002}]{moratz2002qualitative}
Moratz, R.; Nebel, B.; and Freksa, C.
\newblock 2002.
\newblock Qualitative spatial reasoning about relative position.
\newblock In {\em International Conference on Spatial Cognition},  385--400.
\newblock Springer.

\bibitem[\protect\citeauthoryear{Muggleton}{2017}]{muggleton2017meta}
Muggleton, S.~H.
\newblock 2017.
\newblock Meta-interpretive learning: achievements and challenges.
\newblock In {\em International Joint Conference on Rules and Reasoning},
  1--6.
\newblock Springer.

\bibitem[\protect\citeauthoryear{Narayan-Chen \bgroup et al\mbox.\egroup
  }{2017}]{narayan2017towards}
Narayan-Chen, A.; Graber, C.; Das, M.; Islam, M.~R.; Dan, S.; Natarajan, S.;
  Doppa, J.~R.; Hockenmaier, J.; Palmer, M.; and Roth, D.
\newblock 2017.
\newblock Towards problem solving agents that communicate and learn.
\newblock In {\em Proceedings of the First Workshop on Language Grounding for
  Robotics},  95--103.

\bibitem[\protect\citeauthoryear{Peters and
  Schaal}{2008}]{peters2008reinforcement}
Peters, J., and Schaal, S.
\newblock 2008.
\newblock Reinforcement learning of motor skills with policy gradients.
\newblock {\em Neural networks} 21(4):682--697.

\bibitem[\protect\citeauthoryear{Pustejovsky and
  Krishnaswamy}{2016}]{pustejovsky2016LREC}
Pustejovsky, J., and Krishnaswamy, N.
\newblock 2016.
\newblock Voxml: A visualization modeling language.
\newblock {\em Proceedings of LREC}.

\bibitem[\protect\citeauthoryear{Quinlan}{1990}]{quinlan1990learning}
Quinlan, J.~R.
\newblock 1990.
\newblock Learning logical definitions from relations.
\newblock {\em Machine learning} 5(3):239--266.

\bibitem[\protect\citeauthoryear{Randell \bgroup et al\mbox.\egroup
  }{1992}]{randell1992}
Randell, D.; Cui, Z.; Cohn, A.; Nebel, B.; Rich, C.; and Swartout, W.
\newblock 1992.
\newblock {A spatial logic based on regions and connection}.
\newblock In {\em KR'92. Principles of Knowledge Representation and Reasoning:
  Proceedings of the Third International Conference},  165--176.
\newblock San Mateo: Morgan Kaufmann.

\bibitem[\protect\citeauthoryear{Schneider}{2014}]{schneider2014convex}
Schneider, R.
\newblock 2014.
\newblock {\em Convex bodies: the Brunn--Minkowski theory}.
\newblock Number 151. Cambridge university press.

\bibitem[\protect\citeauthoryear{Smart and
  Kaelbling}{2002}]{smart2002effective}
Smart, W.~D., and Kaelbling, L.~P.
\newblock 2002.
\newblock Effective reinforcement learning for mobile robots.
\newblock In {\em Robotics and Automation, 2002. Proceedings. ICRA'02. IEEE
  International Conference on}, volume~4,  3404--3410.
\newblock IEEE.

\bibitem[\protect\citeauthoryear{Smyth}{2007}]{smyth2007case}
Smyth, B.
\newblock 2007.
\newblock Case-based recommendation.
\newblock In {\em The adaptive web}. Springer.
\newblock  342--376.

\bibitem[\protect\citeauthoryear{Veeraraghavan, Papanikolopoulos, and
  Schrater}{2007}]{veeraraghavan2007learning}
Veeraraghavan, H.; Papanikolopoulos, N.; and Schrater, P.
\newblock 2007.
\newblock Learning dynamic event descriptions in image sequences.
\newblock In {\em IEEE Conference on Computer Vision and Pattern Recognition
  (CVPR)},  1--6.
\newblock IEEE.

\bibitem[\protect\citeauthoryear{Williams}{1992}]{williams1992simple}
Williams, R.~J.
\newblock 1992.
\newblock Simple statistical gradient-following algorithms for connectionist
  reinforcement learning.
\newblock {\em Machine learning} 8(3-4):229--256.

\bibitem[\protect\citeauthoryear{Wu \bgroup et al\mbox.\egroup
  }{2015}]{wu2015watch}
Wu, C.; Zhang, J.; Savarese, S.; and Saxena, A.
\newblock 2015.
\newblock Watch-n-patch: Unsupervised understanding of actions and relations.
\newblock In {\em IEEE Conference on Computer Vision and Pattern Recognition
  (CVPR)},  4362--4370.
\newblock IEEE.

\bibitem[\protect\citeauthoryear{Xu \bgroup et al\mbox.\egroup
  }{2014}]{xu2014deep}
Xu, L.; Ren, J.~S.; Liu, C.; and Jia, J.
\newblock 2014.
\newblock Deep convolutional neural network for image deconvolution.
\newblock In {\em Advances in Neural Information Processing Systems},
  1790--1798.

\end{thebibliography}
\bibliographystyle{aaai}

\end{document}